\documentclass[letterpaper, 10 pt, conference]{ieeeconf}  % Comment this line out if you need a4paper

\makeatletter
\let\NAT@parse\undefined
\makeatother
\IEEEoverridecommandlockouts                              % This command is only needed if 
\usepackage{float}  % Add this in your preamble for [H] specifier
\usepackage{graphicx}
\usepackage{subfigure}
\usepackage{caption}
\usepackage{multicol}
\usepackage{multirow}
\usepackage{makecell}
\usepackage{array}
\usepackage[table]{xcolor}
\usepackage{amsmath}
\usepackage{amssymb}
\usepackage{kotex}
\usepackage{cite}   % <- hyperref보다 먼저 로드

\usepackage{hyperref}
\usepackage[normalem]{ulem}

\usepackage{array}
\newcolumntype{C}[1]{>{\centering\arraybackslash}m{#1}}

\newcommand\mycustomsize{\fontsize{6.75}{9}\selectfont} % Define a custom size between footnotesize and scriptsize

\newcommand\peerv[1]{\textcolor{black}{#1}}
\newcommand\pf[1]{\textcolor{black}{#1}}

\newcommand\donguk[1]{\textcolor{black}{#1}}

\newcommand\fn[1]{\textcolor{black}{#1}} % final camera ready
\newcommand\rev[1]{\textcolor{black}{#1}} % final camera ready: revised

\newcommand\ul[1]{\underline{#1}}

\definecolor{myFirst}{HTML}{C9FFB5}
\definecolor{mySecond}{HTML}{E9FFE2}
\definecolor{myThird}{HTML}{FFFFC7}

\newcommand{\rot}[1]{\rotatebox[origin=c]{90}{#1}}

\makeatletter
\newcommand{\SecRef}[1]{Section~\expandafter\@SecRefAux\ref{#1}\@nil}
\def\@SecRefAux#1.#2\@nil{#1.\textit{#2}}
\makeatother
\title{\LARGE \bf
    {AIM}-SLAM: Dense Monocular SLAM via Adaptive and Informative 
    \\ Multi-View Keyframe Prioritization with Foundation Model
}

\author{
    \fn{
    Jinwoo Jeon$^{1}$, 
    Dong-Uk Seo$^{1}$,
    Eungchang Mason Lee$^{2}$,
    and Hyun Myung$^{1*}$,
    }% <-this % stops a space
    \thanks{
        \fn{
        $^{*}$Corresponding author: Prof. Hyun Myung
        }
    }
    \thanks{
        \fn{
        $^{1}$The School of Electrical Engineering, KAIST (Korea Advanced Institute of Science and Technology), Daejeon, 34141, Republic of Korea, 
        {\tt\small \{zinuok, dongukseo, hmyung\}@kaist.ac.kr}
        }
    }
    \thanks{
        \fn{
        $^{2}$KAIST InnoCORE LLM, KAIST, Daejeon, 34141, Republic of Korea, 
        {\tt\small eungchang\_mason@kaist.ac.kr}
        }
    }
}

\begin{document}

\maketitle
\thispagestyle{empty}
\pagestyle{empty}

%%%%%%%%%%%%%%%%%%%%%%%%%%%%%%%%%%%%%%%%%%%%%%%%%%%%%%%%%%%%%%%%%%%%%%%%%%%%%%%%
\begin{abstract}

    % 1) What task is this paper dealing with & why is it important?
    Recent advances in \pf{geometric} foundation models have emerged as a promising alternative for addressing the challenge of dense reconstruction in monocular visual simultaneous localization and mapping~(SLAM). 
    % 2) What are the limitations of existing methodologies?
    Although \pf{geometric} foundation models enable SLAM to leverage variable input views, the previous methods remain confined to two-view pairs or \pf{fixed}-length inputs without sufficient deliberation of geometric context for view selection.
    % 3) What is the methodology proposed in this paper? [main Idea] 
    To tackle this problem, we propose 
    \textit{AIM-SLAM}, a dense monocular SLAM framework that exploits an adaptive and informative multi-view \pf{keyframe prioritization} 
    with dense pointmap predictions from visual geometry grounded transformer~(VGGT). 
    Specifically, we introduce the selective information- and geometric-aware multi-view adaptation~(SIGMA) module, 
    which employs voxel overlap and information gain to retrieve \peerv{a candidate set of keyframes and adaptively determine its size. }
    Furthermore, we formulate a joint multi-view $\mathrm{Sim}(3)$ optimization that enforces consistent alignment across selected views, substantially improving pose estimation accuracy. 
    % 4) How did we prove the effectiveness of the proposed method? 
    The effectiveness of AIM-SLAM is demonstrated on real-world datasets, where it achieves state-of-the-art \rev{pose estimation} performance and \rev{accurate} dense reconstruction \rev{results}. 
    Our system supports ROS integration, with code \fn{is available at \url{https://aimslam.github.io/}}. 

\end{abstract}

%%%%%%%%%%%%%%%%%%%%%%%%%%%%%%%%%%%%%%%%%%%%%%%%%%%%%%%%%%%%%%%%%%%%%%%%%%%%%%%%
\section{Introduction}

% visual SLAM 설명
Visual simultaneous localization and mapping~(SLAM) has traditionally relied on geometric pipelines that exploit handcrafted features and require accurate camera calibration to estimate camera poses~\cite{msckf,svo,orbslam}. 
Recent geometry-aware foundation models such as DUSt3R~\cite{dust3r}, MASt3R~\cite{mast3r}, and VGGT~\cite{vggt} have emerged as compelling alternatives, directly predicting dense 3D pointmaps from uncalibrated RGB inputs.
% , while implicitly accounting for camera intrinsics. 
% Building on these advances, several works have extended foundation models into SLAM, enabling dense reconstruction with uncalibrated monocular images.
Leveraging these advantages, researchers have aimed to extend foundation models into SLAM systems that support dense reconstruction with uncalibrated monocular images~\cite{must3r, mast3r-slam, vggtslam, vggtlong}. 

As the number of input views that foundation models can accommodate has expanded, several approaches have incorporated multi-view reasoning into SLAM~\cite{vggtslam, vggtlong}. To extend to the multi-view setting, most works have followed the sequential design, forming temporal windows of consecutive keyframes and aligning the submaps from each window with optimization. Although this design achieves decent performance, such a simple conjunction of neighboring frames does not fully exploit the potential of multi-view constraints from the foundation model; it often includes redundant frames with limited geometric information gain. While conventional SLAM approaches have explored multi-view keyframe selection methods based on map representation and covisibility~\cite{orbslam, droidslam}, such considerations remain largely unexplored in foundation model-based SLAM, underscoring the need for more principled keyframe prioritization. 

In this context, we propose \textit{AIM-SLAM}, a monocular SLAM framework that leverages an adaptive and informative multi-view prioritization with foundation models. By prioritizing informative views for optimization, rather than relying solely on temporally adjacent frames, our approach \rev{improves} geometric consistency, mitigates scale drift, and enables \rev{consistent} dense reconstruction \rev{across diverse scenes}. 
% As shown in Fig.~\ref{fig:cover}, AIM-SLAM is based on the insight that an appropriately sparse and overlapping subset of past views, selected under informative criteria, provides sufficient geometric constraints for accurate pose and map refinement. 

%%%%%%%%%%%%%%%%%%%%%%
\begin{figure}[t!]
    % \vspace{-5pt}
    \centering
    \includegraphics[width=1.0\columnwidth, keepaspectratio]{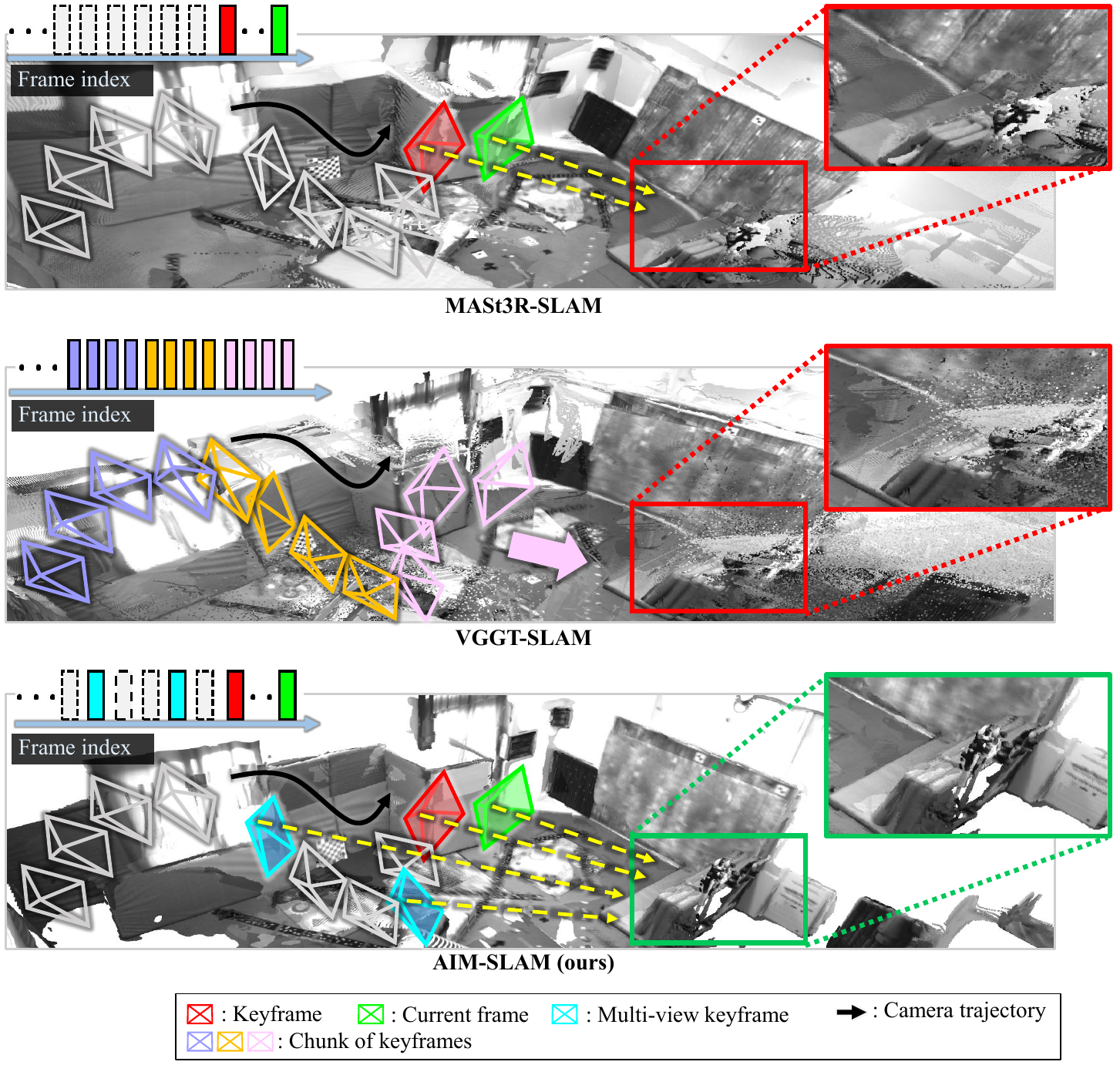}
    \caption{
        Comparison \peerv{among MASt3R-SLAM~\cite{mast3r-slam}, VGGT-SLAM~\cite{vggtslam}}, and the proposed AIM-SLAM. 
        \peerv{MASt3R-SLAM relies on a fixed two-view input, while VGGT-SLAM processes a fixed chunk of consecutive keyframes. 
        In contrast}, AIM-SLAM adaptively prioritizes a variable number of keyframes with high viewpoint overlap and information gain. 
        By jointly optimizing these multi-view inputs in $\mathrm{Sim}(3)$ space, AIM-SLAM achieves accurate and globally consistent dense reconstruction. 
    }
    \label{fig:cover}
    \vspace{-18pt}
\end{figure}
%%%%%%%%%%%%%%%%%%%%%%

As shown in Fig.~\ref{fig:cover}, AIM-SLAM adaptively {prioritizes} candidate keyframes that maximize 3D scene overlap and information gain. 
A subsequent stability \pf{criterion} regulates the incorporation of these views into optimization, and the prioritization process is repeated until convergence. Through this adaptive mechanism, AIM-SLAM leverages foundation models in a manner \fn{suited to} SLAM, enabling both accurate and consistent reconstruction. 

% outro
% key claims 
In summary, our key contributions are as follows: 
\begin{itemize}
    % Adaptive Multi-view Tracking: 
    \item
        An adaptive and informative multi-view \pf{prioritization} is introduced to construct a sparse yet highly overlapping keyframe set. 
        To this end, we propose the selective information- and geometric-aware multi-view adaptation~(SIGMA) module, while a stability \pf{criterion} adaptively regulates their incorporation into the frontend visual odometry, ensuring geometric consistency and minimizing redundancy in foundation model-based SLAM. 
    % Multi-view Sim(3) Optimization:
    \item 
        We present a joint multi-view $\mathrm{Sim}(3)$ optimization in foundation model-based SLAM, enabling accurate alignment across multiple views without requiring camera calibration. 
    % effect
    \item 
        \rev{
        Extensive evaluations on public datasets validate the effectiveness of AIM-SLAM in pose estimation and dense reconstruction. } 
        The code \fn{is} publicly released, with ROS integration also provided. 
        % to facilitate deployment in robotic applications. 
\end{itemize}

% 6) 논문 구성
%-------------------------------------------------------------------------
% HT comment: 꼭 필요한가요? 살짝 redundant한 듯?!
% The rest of this paper is organized as follows: Section \ref{sec:related} reviews related studies. Section \ref{sec:proposed} describes the proposed method in detail and Section \ref{sec:exp_result} presents a detailed analysis of the experimental results. Finally, Section \ref{sec:cons} summarizes our contributions as well as future works.

\section{Related Works} \label{sec:related}

%-------------------------------------------------------------------------
\subsection{Classical and Learning-\pf{based} Visual SLAM}\label{sec:rw-classical}

%------------------------------
% \textbf{Conventional approach}~
Conventional visual SLAM pipelines could be categorized by their input-processing strategies. 
Indirect methods estimate motion by minimizing reprojection error of sparse salient \fn{primitives, such as points~\cite{vinsmono, orbslam, orbslam3} and line segments~\cite{avoiding, plfvins, uvslam}. }
Direct methods minimize photometric error over pixel intensities, yielding (semi-)dense maps from monocular~\cite{lsdslam, dtam, dso}, stereo~\cite{zhang2021stereo, stereodso}, or RGB-D inputs~\cite{kerl2013dense, elasticfusion}. 
Semi-direct approaches combine both paradigms~\cite{svo, forster2016svo}. 
While these approaches have enabled decades of progress, such methods remain constrained by handcrafted modules and reliance on accurate calibration in monocular settings.

%------------------------------
% \textbf{Learning-based approach}~
With the advent of deep learning~\cite{cnn}, these handcrafted modules have increasingly been replaced by learned counterparts. This includes data-driven descriptors and matchers for 
robust feature tracking~\cite{superpoint,superglue}, 
end-to-end networks regressing pose (and often depth) from RGB video~\cite{sfmlearner,deepvo,undeepvo,iyer2018geometric}, 
and hybrid methods injecting learned priors such as depth~\cite{codeslam, d3vo}, semantics~\cite{cnnslam}, or optical flow~\cite{d3vo,zhan2020visual} into geometric back-ends.  
% \textbf{Advanced Learning-based Visual SLAM}~
More recently, differentiable dense bundle adjustment~(DBA) has been explored in learning-based SLAM: 
DROID-SLAM~\cite{droidslam} coupled ConvGRU updates with DBA, DPVO~\cite{dpvo} introduced a path-based formulation, and GO-SLAM~\cite{goslam} extended DBA with full bundle adjustment and an implicit surface model. 

% However, all these conventional methods assume known intrinsics for reprojection and primarily optimize in $\mathrm{SE}(3)$ space. 
% While some monocular systems introduce $\mathrm{Sim}(3)$ constraints to mitigate scale drift, these formulations still rely on calibrated cameras. 
% In contrast, geometry-aware foundation models infer both depth and intrinsics directly from uncalibrated images, eliminating this dependency and opening the door to more general formulations. 
% Building on this capability, we formulate a joint multi-view $\mathrm{Sim}(3)$ optimization tailored for foundation model-based SLAM, enabling accurate alignment across multiple overlapping views under uncalibrated conditions. 

% However, conventional methods rely on known intrinsics for reprojection and primarily optimize in $\mathrm{SE}(3)$ space. 
% Although some monocular systems adopt $\mathrm{Sim}(3)$ constraints to mitigate scale drift, they still depend on calibrated cameras. 
% In contrast, geometry-aware foundation models infer both depth and intrinsics directly from uncalibrated images, removing this dependency. Building on this capability, we present, to the best of our knowledge, the first joint multi-view $\mathrm{Sim}(3)$ optimization in foundation model-based SLAM, enabling accurate alignment across multiple overlapping views under uncalibrated conditions. 

%-------------------------------------------------------------------------
\subsection{\pf{Visual} Foundation Models for 3D Geometry}\label{sec:rw-foundation}

%------------------------------
% \textbf{Two-view models}~
% dust3r

Recent geometry-aware foundation models have demonstrated strong ability to infer dense 3D structure directly from uncalibrated images. 
DUSt3R~\cite{dust3r} first demonstrated this from image pairs, and subsequent works extended this principle to multi-view inference via learnable recurrent memory~\cite{cut3r,spann3r}, all-to-all attention~\cite{fast3r}, or additional priors such as intrinsics and depth~\cite{pow3r}. 
For dynamic scenes, MonST3R~\cite{monst3r} leveraged optical flow, while Easi3R~\cite{easi3r} decomposed cross-attention maps to separate moving from static geometry. 
% mast3r
Among these, MASt3R~\cite{mast3r} introduced per-pixel descriptors that enabled downstream systems for structure-from-motion~\cite{mast3r-sfm} and real-time dense SLAM~\cite{mast3r-slam}. 
In a parallel line of work, VGGT~\cite{vggt} generalized to arbitrary multi-view inputs, jointly predicting intrinsics, poses, depth, tracks, and dense pointmaps in a single feed-forward pass.

% SLAM applications
These foundation models have recently been adapted for SLAM. MASt3R-SLAM~\cite{mast3r-slam} demonstrated the first real-time dense monocular SLAM leveraging a reconstruction prior, but its reliance on the adjacent two-view input restricts parallax diversity and can lead to structural inconsistencies under challenging motions. 
VGGT-SLAM~\cite{vggtslam} extended VGGT to online settings by batching $16$–$32$ consecutive frames into submaps aligned via $\mathrm{SL}(4)$ optimization, while VGGT-Long~\cite{vggtlong} scaled this to larger $60$–$75$ frame windows with $\mathrm{Sim}(3)$ refinement. 
Although these submap-based methods enable online SLAM with foundation models, they (i) mainly rely on the $N$ most adjacent views, which often contain redundant overlap even when keyframes are selected, and (ii) require large, fixed window sizes to ensure sufficient geometric coverage, treating SLAM as deferred submap registration rather than continuous multi-view tracking.

To tackle these problems, we propose an overlap-aware keyframe prioritization method formulated in an adaptive and informative manner, preserving VGGT’s geometric fidelity while avoiding redundant inference and retaining only informative views. 
As a result, the proposed approach offers a more scalable solution compared with fixed window foundation model SLAM systems.

\section{AIM-SLAM} \label{sec:proposed}

%%%%%%%%%%%%%%%%%%%%%%
\begin{figure*}[t!]
    \centering
    \includegraphics[width=1.95\columnwidth, keepaspectratio]{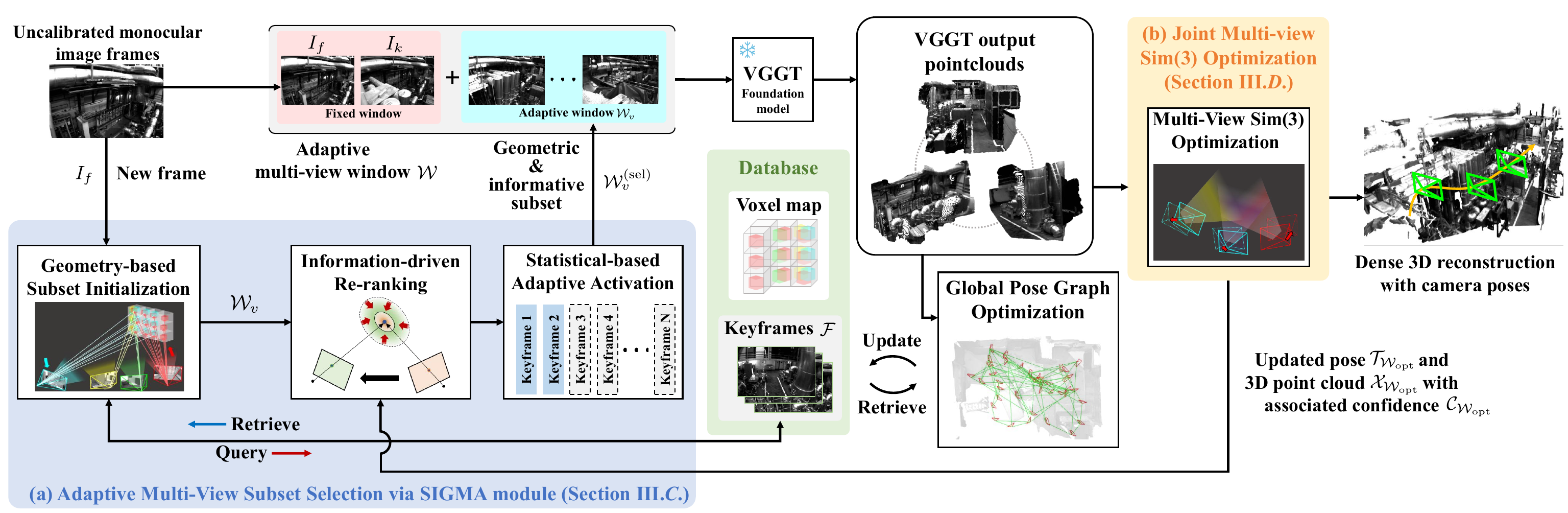}
    \caption{
        Overall architecture of AIM-SLAM. 
        The frontend consists of 
        (a)~multi-view prioritization method via the proposed SIGMA module, followed by VGGT-based dense pointmap inference, and 
        (b)~joint multi-view $\mathrm{Sim}(3)$ optimization to mitigate short- and mid-term drift. 
        The backend loop closure module performs global pose-graph optimization to ensure global consistency. 
    }  
    \label{fig:overall-arch}
\vspace{-0.4cm}
\end{figure*}  
%%%%%%%%%%%%%%%%%%%%%%

Fig.~\ref{fig:overall-arch} shows the overall framework of AIM-SLAM, 
which consists of
(a)~adaptive and informative multi-view keyframe \pf{prioritization} with VGGT inference and
(b)~joint multi-view $\mathrm{Sim}(3)$ pose optimization for frontend visual odometry, while loop closure with global pose-graph optimization runs asynchronously in a separate backend thread.

%-------------------------------------------------------------------------
\subsection{Preliminaries: VGGT}

% summary & encoder
VGGT~\cite{vggt} is a feed-forward network that processes an arbitrary-length sequence~$\mathcal{I} = \{I_i,\dots,I_{i+N}\}$ using DINOv2~\cite{dino} patch embeddings with alternating local (within-frame) and global (across-frame) self-attention layers. 
% decoder
Its decoder~\cite{dpt} predicts per-frame depth with confidence, 3D points, correspondences, and camera parameters.  
% what do we use?
In our pipeline, each depth map is backprojected into a point cloud expressed in the first frame’s coordinates, with its associated confidence, as this yields higher accuracy than directly using the raw 3D points~\cite{vggt}. 
We denote the point clouds and confidences as $\mathcal{X} = \{\mathbf{X}_{i|i}, \cdots, \mathbf{X}_{i|i+N}\}$ and $\mathcal{C} = \{\mathbf{C}_{i|i}, \cdots, \mathbf{C}_{i|i+N}\}$, respectively.

%-------------------------------------------------------------------------
\subsection{Problem Definition}

Our goal is to estimate globally consistent camera poses and dense 3D reconstructions from uncalibrated monocular images. 
Given an image stream~$\{I_1, I_2, \dots\}$, the absolute pose of $j$-th frame is denoted as $\mathbf{T}^w_j$, which transforms points from camera frame $j$ into the world frame $w$. 
A relative transformation from frame~$i$ to frame~$j$ is defined as $\mathbf{T}^{i}_j = (\mathbf{T}^{w}_i)^{-1}\,  \mathbf{T}^{w}_j$, belonging to $\mathrm{Sim}(3)$ and comprising scale~$s \in \mathbb{R}^{+}$, rotation~$\mathbf{R} \in \text{SO}(3)$, and translation~$\mathbf{t} \in \mathbb{R}^3$. 
For a given frame pair~$(i,j)$, $\mathbf{X}_{i \mid j}$ and $\mathbf{C}_{i\mid j}$ denote the 3D pointmap and its per-point confidence, predicted by VGGT from frame~$j$ and expressed in $i$’s coordinates. 

% intro
AIM-SLAM is a keyframe-based tracking framework following MASt3R-SLAM~\cite{mast3r-slam}, but advances beyond fixed-window designs by introducing 
(i)~adaptive, informative and overlap-aware multi-view \pf{prioritization} \peerv{regulated} by VGGT predictions, and 
(ii)~a joint $\mathrm{Sim}(3)$ optimization that ensures scalable and consistent reconstruction.

%%%%%%%%%%%%%%%%%%%%%%
\begin{figure}[t]
% \begin{figure}[h!]
    % \vspace{-5pt}
    \centering
    \includegraphics[width=0.65\columnwidth, keepaspectratio]{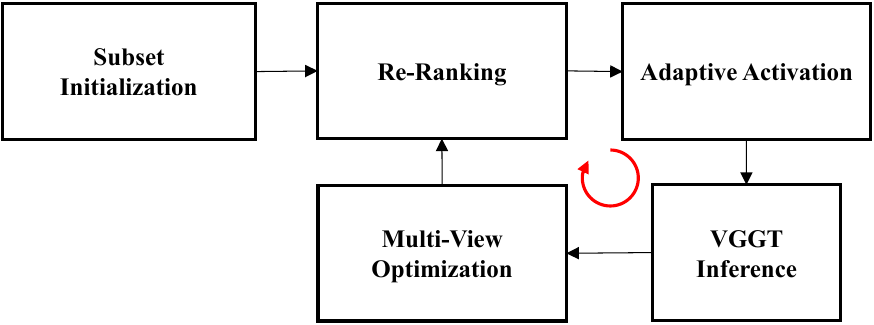}
    \caption{
        Block diagram of the proposed SIGMA module, which consists of three stages: (i)~geometry-based subset initialization via voxel overlap, 
        (ii)~information-driven re-ranking based on covariance reduction, and 
        (iii)~adaptive activation \peerv{regulated} by a stability test. 
        After each multi-view optimization, updated poses and confidences recurrently trigger the re-ranking process. 
    }
    \label{fig:sigma_framework}
    % \vspace{-35pt}
% \end{figure} 
\vspace{-0.5cm}
% \vspace{\vsfig}
\end{figure}
%%%%%%%%%%%%%%%%%%%%%%

%-------------------------------------------------------------------------
\subsection{Adaptive Multi-view \pf{Prioritization} via SIGMA \pf{Module}} \label{sec:}
% $\mathcal{W}_v \subset \{I_f\} \cup \mathcal{F}$

% intro
Leveraging VGGT’s ability to process an arbitrary number of views, AIM-SLAM adaptively constructs a sparse yet highly overlapping and informative keyframe subset that serves as the VGGT input. 
To this end, we define the candidate set~$\mathcal{W}_v$ for the input subset~$\mathcal{W}$.

% 구체화
To construct~$\mathcal{W}_v$, we propose the SIGMA module, which consists of three stages: 
(a)~geometry-based initialization of candidate views using voxel-overlap scores, 
(b)~information-driven re-ranking of these candidates based on covariance reduction, and 
(c)~adaptive subset activation \peerv{regulated} by a statistical stability test. 
The \peerv{overall procedure} of the proposed SIGMA module is summarized in Fig.~\ref{fig:sigma_framework}. 

During the operation of the SIGMA module, VGGT input subset is denoted as $\mathcal{W} = \mathcal{W}_0 \cup \mathcal{W}_v^{(\mathrm{sel})}$, where $\mathcal{W}_0 = \{ I_f, I_k, I_b \}$ contains the current frame~$I_f$, the last keyframe~$I_k$, and the best candidate~$I_b \in \mathcal{W}_v$\peerv{;} $\mathcal{W}_v^{(\mathrm{sel})} \subseteq \mathcal{W}_v$ denotes the adaptively activated subset beyond \peerv{the default triplet~$\mathcal{W}_0$}. 
Subsequently, $\mathcal{W}$ is fed into VGGT for multi-view inference.

The SIGMA module performs subset initialization and re-ranking with respect to the last keyframe~$I_k$ rather than the incoming frame~$I_f$. This is because $I_f$ and $I_k$ inherently maintain overlap, and $I_f$ is promoted to a new keyframe once this overlap falls below a threshold.
In addition, because keyframe pointmaps are fused via confidence-weighted averaging~(Section~\ref{sec:joint_opt}), $I_k$ serves as a stable anchor for accumulating reliable multi-view information.

% 1) voxel map 기반 overlap keyframe candidate 선정
\subsubsection{Geometry-\pf{based} Initial Subset Construction}

% Once the new keyframe~$I_k$ is determined, the subset is first initialized as $\mathcal{W} = \{I_k\}$. 

\begin{figure}[t]
% \begin{figure}[h!]
    % \vspace{-5pt}
    \centering
    \includegraphics[width=0.725\columnwidth, keepaspectratio]{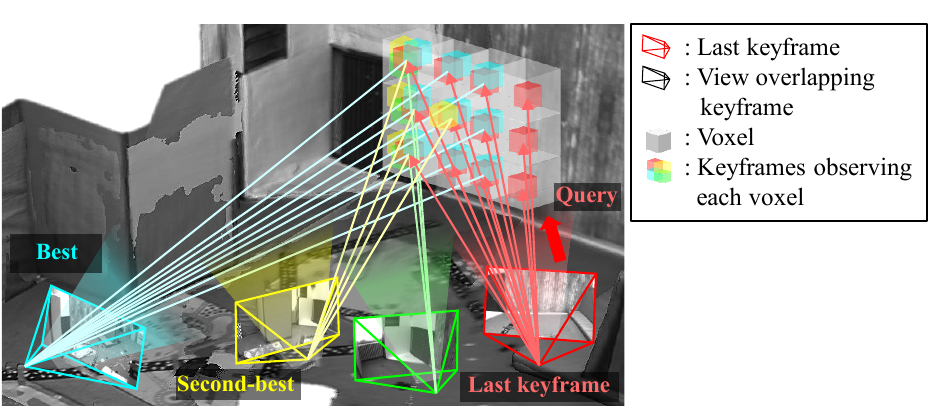}
    \caption{
        Example of a voxel-indexed keyframe map for computing view overlap. Each voxel stores the IDs of keyframes that observe it. 
        Using the last keyframe as the query, the system counts \fn{shared} voxels and selects the $\text{top-}N$ overlapping keyframes to \peerv{initialize} the multi-view subset. 
    }
    \label{fig:voxelmap}
    \vspace{-0.25cm}
% \end{figure} 
\vspace{-0.5cm}
\end{figure}

We propose a voxel-indexed keyframe map (Fig.~\ref{fig:voxelmap}), where each voxel stores the IDs of keyframes that observe it. 
For the last keyframe~$I_k$, we compute the voxel-overlap score as follows: 
%%%%%%%%%%%%%%%%%%%%%%%%%%%
\begin{equation}
\begin{aligned}
    O(I_k, I_i) 
        = \left|\, v(I_k) \cap v(I_i) \,\right|
    , \ 
    I_i \in \mathcal{F} \setminus \{I_k\},
\label{}
\end{aligned}
\end{equation}
%%%%%%%%%%%%%%%%%%%%%%%%%%%
where $O(\cdot, \cdot)$ denotes the overlap score, $v(I)$ denotes the set of voxels observed by $I$, and $I_i$ denotes \peerv{a} keyframe in \peerv{the} keyframe set~$\mathcal{F} \setminus \{I_k\}$. 
The $\text{top-}N$ keyframes by this score form the initial candidate set~$\mathcal{W}_v$. 
As the 3D points predicted by the foundation model is highly dense, voxel-wise associations are used instead of raw points to provide a more compact and efficient representation of co-visibility.

% novelty
In contrast to prior voxel maps that index 3D landmarks for point retrieval~\cite{muglikar2020voxel,yuan2025voxel}, our map explicitly records keyframe visibility, shifting the role of voxelization from point-level data association to view-level selection, thereby aligning with our primary objective of adaptively constructing a multi-view input tailored for foundation model inference.

% 2) cov reduction 기반 re-rank
\subsubsection{Information-\pf{driven} Subset Re-\pf{ranking}}

The voxel-overlap candidates ensure sufficient co-visibility, but geometric overlap alone does not reflect the informativeness of each view. 
To \pf{prioritize} the candidate keyframes, we re-rank $\mathcal{W}_v$ using information criterion based on the reduction of 3D point covariances, assuming that 3D points predicted by the foundation model follow Gaussian distribution. 
Because the last keyframe is typically the least optimized than other keyframes yet has the strongest influence on the current frame, we adopt a strategy that prioritizes candidate views to maximize the information gain of the point cloud of the last keyframe.

Formally, let $\mathbf{P}_{k}^{-}(\mathbf{x}_k)$ denote the prior covariance of a 3D point $\mathbf{x}_k$ observed in the last keyframe $I_k$, derived from pixel noise and fused confidence aggregated in the last keyframe under a Gaussian assumption. 
For brevity, we write $\mathbf{P}_k = \mathbf{P}(\mathbf{x}_k)$. 
Following the standard extended Kalman filter update form~\cite{welch1995introduction, das2015entropy}, incorporating a candidate view $I_j \in \mathcal{W}_v$ updates the prior covariance as follows: 
%%%%%%%%%%%%%%%%%%%%%%%%%%%
\begin{equation}
\begin{aligned}
   \mathbf{P}_{k \rightarrow j}^{+}
        = 
        \mathbf{P}_{k}^{-}
        -
        \mathbf{P}_{k}^{-} \mathbf{J}_r^\top 
        \bigl( \mathbf{R} + \mathbf{J}_r \mathbf{P}_{k}^{-} \mathbf{J}_r^\top \bigr)^{-1}
        \mathbf{J}_r\,\mathbf{P}_{k}^{-},
\label{}
\end{aligned}
\end{equation}
%%%%%%%%%%%%%%%%%%%%%%%%%%%
where $\mathbf{P}_{k \rightarrow j}^{+}$ denotes the posterior covariance after adding $I_j$, $\mathbf{J}_r$ denotes the Jacobian of the ray-based residual~\cite{mast3r-slam} when reprojecting the 3D points from the last keyframe~$I_k$ into the candidate view~$I_j$, and $\mathbf{R}$ denotes the measurement covariance. 
\peerv{Unlike} prior entropy-based methods~\cite{das2015entropy}, which assume 2D image-space noise, we leverage foundation-model predictions to define $\mathbf{R}$ as a full $3 \times 3$ covariance, obtained by propagating pixel noise through the inverse projection Jacobian (i.e., the Jacobian of the pinhole inverse projection function) with VGGT-predicted depth confidence.  
For efficiency, \peerv{only a subset of points with low confidence is considered in this computation. }

%%%%%%%%%%%%%%%%%%%%%%
%%%%%%%%%%%%%%%%%%%%%%

% 
Subsequently, the information gain of view $I_j$ relative to keyframe $I_k$ is then quantified as the following entropy reduction: 
%%%%%%%%%%%%%%%%%%%%%%%%%%%
\begin{equation}
\begin{aligned}
    \Gamma(I_k, I_j) 
        = 
        \!\!\sum_{\mathbf{x}_k \in \Omega_{k \rightarrow j}}
        \!\!\frac{1}{2}
        \log \frac{\det(\mathbf{P}_{k}^{-})}{\det(\mathbf{P}_{k \rightarrow j}^{+})},
\label{}
\end{aligned}
\end{equation}
%%%%%%%%%%%%%%%%%%%%%%%%%%%
where $\Gamma(\cdot, \cdot)$ denotes the information gain score, $\Omega_{k \rightarrow j}$ denotes the valid point set after reprojection, and $\det(\cdot)$ denotes the determinant. 
Finally, the candidate set $\mathcal{W}_v$ is re-ranked by $\Gamma$ to form the ordered subset. 

\begin{figure}[t]
% \begin{figure}[h!]
    % \vspace{-5pt}
    \centering
    \includegraphics[width=0.60\columnwidth, keepaspectratio]{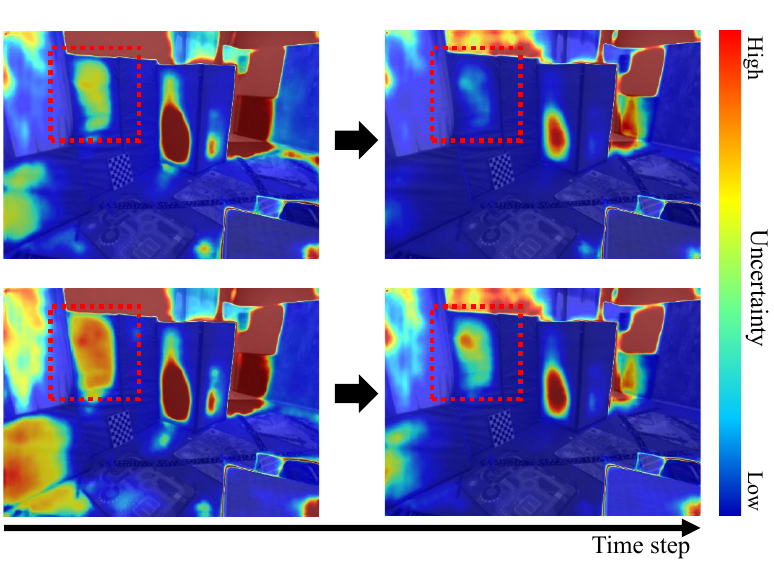}
    \caption{
        Effect of the SIGMA module on keyframe uncertainty reduction. Compared with the case without re-ranking (lower), incorporating information-driven re-ranking (upper) significantly decreases keyframe uncertainty, computed as the inverse of the fused point confidence aggregated across observations during optimization. 
        This \fn{shows} that the SIGMA module effectively retrieves informative frames to refine the keyframe. 
        Uncertainty is visualized in color, with higher values shown in warmer colors. 
        Regions with pronounced differences are highlighted with red rectangles. 
    }
    \label{fig:sigma_cov_reduct}
    % \vspace{-35pt}
% \end{figure} 
\vspace{-0.6cm}
% \vspace{\vsfig}
\end{figure}

The SIGMA module effectively balances geometric co-visibility and information gain, yielding an ordered subset well-suited for multi-view inference.
As shown in Fig.~\ref{fig:sigma_cov_reduct}, the keyframes prioritized through the re-ranking stage of the SIGMA module significantly reduce the covariance of the last keyframe compared with the case without re-ranking, confirming that our strategy selects informative frames. 

%We re-rank $\mathcal{W}_v$ after each optimization convergence for the current frame to reflect updated depth confidence.
After multi-view optimization is performed \peerv{on the window selected by the SIGMA module}, the re-ranking is recurrently performed to \peerv{reflect} the effect of the updated depth confidence by the optimization. 
To avoid oscillations in the adaptive activation process, frames already in the input subset~$\mathcal{W}$ are excluded from re-ranking in subsequent updates. 
Only remaining candidates are reordered, preventing an activated view from being repeatedly swapped with alternatives.

\subsubsection{Adaptive Subset Activation with Stability \pf{Criterion}}

After the re-ranking stage, the candidate subset~$\mathcal{W}_v$ contains diverse informative views, but it is not necessary to activate all of them. 
In practice, a smaller number of views is often sufficient, \donguk{so we assess the statistical stability to determine whether candidate keyframes should be activated. As a result,} only a subset of $\mathcal{W}_v$ is activated as needed, yielding a compact yet effective input~$\mathcal{W}$ to VGGT. 
Adaptive activation starts from the default three-view baseline~$\mathcal{W}_0$, and additional \donguk{candidate keyframe} from $\mathcal{W}_v$ \donguk{is} appended iteratively based on the statistical stability of the optimization. 

%so only a subset of $\mathcal{W}_v$ is activated as needed, yielding a compact yet effective input~$\mathcal{W}$ to VGGT. 

%%%%%%%%%%%%%%%%%%%%%%
%%%%%%%%%%%%%%%%%%%%%%

To quantify such stability, we employ the reduced \pf{Chi-square} test, a standard goodness-of-fit metric in weighted least squares. 
In previous SLAM methods, the test has been applied for outlier rejection~\cite{chisqaure_outlier1, chisqaure_outlier2} \donguk{or} for directly assessing the statistical stability of the optimization itself~\cite{chisqaure_opt1}. We focus on the latter purpose to \peerv{regulate} the adaptive expansion of the keyframe set. 
Formally, let $\mathbf{b}_0 \in \mathbb{R}^M$ be the whitened residual vector, obtained by normalizing the original residuals, which will be detailed in the following section, and $\mathbf{A}_0 \in \mathbb{R}^{M \times p}$ the corresponding Jacobian at the current linearization. 
Under Gaussian noise assumptions, the residual sum of squares follows a \pf{Chi-square} distribution with $\nu=M-\mathrm{rank}(\mathbf{A}_0)$ degrees of freedom~\cite{chisquare_verified}, leading to the reduced statistic as follows: 
%%%%%%%%%%%%%%%%%%%%%%%%%%%
\begin{equation}
\begin{aligned}
    \varkappa \;=\; \frac{\mathbf{b}_0^\top \mathbf{b}_0}{M-\mathrm{rank}(\mathbf{A}_0)} \;\sim\; 
        \chi^2 (\,M-\mathrm{rank}(\mathbf{A}_0)),
\label{}
\end{aligned}
\end{equation}
%%%%%%%%%%%%%%%%%%%%%%%%%%%
where $\varkappa$ denotes the reduced \pf{Chi-square} statistic. 
If $\varkappa \le 1.0$, the configuration is considered stable and the window remains at the default three views. 
If $\varkappa > 1.0$, we iteratively append \donguk{an} additional keyframe $I_v \in \mathcal{W}_v$ in order and re-evaluate $\varkappa$ after \donguk{multi-view} optimization. 
When the inclusion of an additional \donguk{keyframe} results in a decrease of $\varkappa$, indicating improved stability, the \donguk{keyframe} is retained and further expansion is attempted. 
Conversely, if $\varkappa$ increases, the extension is \donguk{considered} unhelpful and the window reverts to the default three-view configuration.

% 3) 최종 합치기
\donguk{After the assessment of statistical stability, the final input for VGGT} is \donguk{then} defined as 
$
\mathcal{W} = \mathcal{W}_0 \cup \mathcal{W}_v^{(\mathrm{sel})}
$, 
where $\mathcal{W}_v^{(\mathrm{sel})} \subseteq \mathcal{W}_v$ denotes the adaptively activated subset beyond the default three views. 
% Multi-view pointmap inference
%We then feed $\mathcal{W}$ into VGGT to obtain per-frame 
%point clouds
\donguk{From the VGGT inference, we obtain per-frame point clouds}
~$\mathcal{X}_\mathcal{W}
=
\{ 
    \mathbf{X}_{f \mid i} \mid I_i \in \mathcal{W}
\}
$, 
confidences~$\mathcal{C}_\mathcal{W}
=
\{ 
    \mathbf{C}_{f \mid i} \mid I_i \in \mathcal{W}
\}$, 
and intrinsics~$\mathcal{K}_\mathcal{W}
= 
\{ 
    \mathbf{K}_{i} \mid I_i \in \mathcal{W}
\}$, 
which are subsequently used in the joint multi-view optimization described in the following section.

%----------------------------------
\subsection{Joint \pf{Multi-view} \(\mathrm{Sim}(3)\) Optimization with Hybrid Residual}\label{sec:joint_opt}

%----------------------------------
\subsubsection{Tracking and Keyframe Management}

To perform multi-view optimization, we establish correspondences between keyframes in the subset~$\mathcal{W}$ using the ray-matching strategy of MASt3R-SLAM, extended here to handle multiple views. 
Given dense pointmaps predicted by VGGT, each pixel $\mathbf{p}_i$ in frame $I_i$ corresponds to a unique 3D point $\mathbf{x}_{f \mid i}$. 
Correspondences with another frame $I_j$ are then obtained by minimizing the angular difference between their unit rays. 
This ray-based formulation provides scale-invariant dense matches, mitigating VGGT’s mild scale inconsistency while avoiding the overhead of using raw correspondences directly estimated by VGGT across all views. 
If the ratio of valid correspondences in the current frame $I_f$ falls below a threshold, $I_f$ is promoted to a new keyframe and added to~$\mathcal{F}$.

\subsubsection{Optimization}

% problem formulation
For optimization, the keyframes in~$\mathcal{W}$ are arranged in temporal order so that each relative transformation~$\mathbf{T}^{i}_{j} \in \mathrm{Sim}(3)$ naturally represents the motion from an earlier frame~$I_i$ to a later frame~$I_j$. 
To this end, we define an optimization-ordered subset~$\mathcal{W}_\text{opt}$ as the reverse of~$\mathcal{W}$ as  
$
    \mathcal{W}_\text{opt} = \{ I_m, \dots, I_k, I_f \}
$, where $I_m$ denotes the oldest keyframe in the subset and $I_f$ the current frame.
The state vector of adjacent relative transformations is then defined as 
$
\mathcal{T}_{\mathcal{W}_\text{opt}}
    = 
    \begin{bmatrix}
        \mathbf{T}^{m}_{v}, & \cdots, & \mathbf{T}^{k}_{f}
    \end{bmatrix}
$.

% residual
For each \donguk{ordered} adjacent frame pair $(I_i, I_j) \in \mathcal{W}_{\mathrm{opt}}$, we combine ray-based and pixel-based reprojection terms~\cite{mast3r-slam} to define the residual as follows: 
%%%%%%%%%%%%%%%%%%%%%%%%%%%
\begin{equation}
\begin{aligned}
    % \mathbf{r}_{ij} = 
    %     \Psi(\mathbf{X}_{j|j}) 
    %     - 
    %     \Psi(\mathbf{T}^{i}_{j}\, \mathbf{X}_{i|i}).  
    \mathbf{r}_{ij} 
        &=
        \left(
            \Psi_\text{ray}(\mathbf{X}_{i|i}) 
            - 
            \Psi_\text{ray}(\mathbf{T}^{i}_{j}\, \mathbf{X}_{j|j}) 
        \right)
        \\
        &+
        \left(
            \Psi_\pi(\mathbf{K}_i, \mathbf{X}_{i|i}) 
            - 
            \Psi_\pi(\mathbf{K}_i, \mathbf{T}^{i}_{j}\, \mathbf{X}_{j|j})
        \right), 
\label{}
\end{aligned}
\end{equation}
%%%%%%%%%%%%%%%%%%%%%%%%%%%
where
$\mathbf{r}_{ij}$ denotes the reprojection residual, 
$\Psi_\text{ray}(\cdot)$ denotes the ray-normalization function that projects a 3D point onto the unit sphere, 
and $\Psi_\pi(\cdot)$ denotes the pinhole camera projection with VGGT estimated intrinsics $\mathbf{K}_i$. 
As VGGT-predicted intrinsics are not perfectly calibrated, we do not enforce a single global set; instead, each pair adopts the intrinsics of its preceding keyframe $I_i$, whose estimates are more stable due to repeated averaging. 
The \donguk{total} joint multi-view residual \donguk{$\mathbf r$} is then defined as the weighted sum of all pairwise residuals as follows: 
%%%%%%%%%%%%%%%%%%%%%%%%%%%
\begin{equation}
\begin{aligned}
    \mathbf{r}
    = 
    \!\!\!\!\!\!\sum_{(I_i, I_j) \in \mathcal{W}_\text{opt}}
       \!\!\!\! \mathbf{r}_{ij} 
    = \!\!\!\!\!\!\sum_{(I_i, I_j) \in \mathcal{W}_\text{opt}}
    \!\!\left(
        \sum_{(a, b) \in \mathbf{m}_{i \rightarrow j}}
            \left(
                \frac{\mathbf{r}_{ij}^{ab}}
                     {^{ab}w_{ij}}
            \right)_\rho
    \right),
\label{eq:multi_view_residual}
\end{aligned}
\end{equation}
%%%%%%%%%%%%%%%%%%%%%%%%%%%
where 
$\mathbf{m}_{i \rightarrow j}$ denotes the set of pixel correspondences from frame~$i$ to frame~$j$; 
$\mathbf{r}_{ij}^{ab}$ denotes the hybrid residual defined for each correspondence pair and $(\cdot)_\rho$ denotes the Huber norm.  
$^{ab}w_{ij}$ denotes the per-point residual weight~\cite{mast3r-sfm,mast3r-slam}, respectively. 
Following MASt3R-SfM~\cite{mast3r-sfm} and MASt3R-SLAM~\cite{mast3r-slam}, $^{ab}w_{ij}$ is defined as the geometric mean of per-frame confidences. In our formulation, the matching confidences are replaced with per-pixel confidences predicted by VGGT.

% jacobian
Finally, the left Jacobian of each residual with respect to the Lie algebra perturbation $\boldsymbol{\tau} \in \mathfrak{sim}(3)$ is computed and stacked into the global Jacobian as follows: 
%%%%%%%%%%%%%%%%%%%%%%%%%%%
\begin{equation}
\begin{aligned}
    \mathbf{J}
    = 
    \begin{bmatrix}
        \mathbf{J}_{mv},
        &
        \cdots,
        &
        \mathbf{J}_{kf}
    \end{bmatrix}
    = 
    \begin{bmatrix}
        \frac{\partial \mathbf{r}}{\partial \boldsymbol\tau^{m}_{v}},
        &
        \cdots,
        &
        \frac{\partial \mathbf{r}}{\partial \boldsymbol\tau^{k}_{f}}
    \end{bmatrix}
    . 
\label{}
\end{aligned}
\end{equation}
%%%%%%%%%%%%%%%%%%%%%%%%%%%

The optimization problem is formulated as a Levenberg–Marquardt scheme with an iteratively reweighted least squares~(IRLS) solver. 
We construct the Hessian matrix from the Jacobian and information weights, and solve for the pose update vector 
$\boldsymbol{\tau}_{\mathcal{W}_\text{opt}}$ to update the state as 
$
\mathcal{T}_{\mathcal{W}_\text{opt}} \leftarrow 
        \boldsymbol\tau_{\mathcal{W}_\text{opt}} \oplus \mathcal{T}_{\mathcal{W}_\text{opt}}
$.

The operator~$\oplus$ is the left-plus operator, which updates the exponential map~$\mathrm{Exp}(\cdot)$ from~$\mathfrak{sim}(3)$ to~$\mathrm{Sim}(3)$ to update the entire state vector, and is formally defined as follows: 
%%%%%%%%%%%%%%%%%%%%%%%%%%%
\begin{equation}
\begin{aligned}
    \boldsymbol\tau_{\mathcal{W}_\text{opt}} \oplus \mathcal{T}_{\mathcal{W}_\text{opt}}
    \triangleq 
    \begin{bmatrix}
        \mathrm{Exp}(\boldsymbol\tau^{m}_{v}) \circ \mathbf{T}^{m}_{v},
        \cdots,
        \mathrm{Exp}(\boldsymbol\tau^{k}_{f}) \circ \mathbf{T}^{k}_{f}
    \end{bmatrix}. 
\label{}
\end{aligned}
\end{equation}
%%%%%%%%%%%%%%%%%%%%%%%%%%%
% pose accumulation 
To resolve the \emph{gauge freedom} when converting relative poses into the world frame, the earliest frame $I_m$ is fixed. Subsequently, keyframe pointmaps are fused via confidence-weighted averaging~\cite{mast3r-slam}, and VGGT-predicted focal lengths are recursively averaged (principal point assumed at the image center~\cite{vggt})  
% \redsout{These designs mitigating noise and scale drift, ensuring that $\mathrm{Sim}(3)$-based alignment remains robust without resorting to higher-DOF transformations.}

% Unlike prior foundation model-based SLAM systems that rely on external descriptor extractors or retrieval-specific models~\cite{vggtslam,mast3r-slam,salad}, 

% intro
On the other hand, to reduce long-term drift, we implement loop closure by reusing the first-layer token~$z_k$ from VGGT. 
These DINOv2-based patch embeddings have proven effective for visual recognition and suffice as lightweight global descriptors~\cite{dino, keetha2023anyloc}.  
Loop candidates are retrieved via cosine similarity against the stored token database, with the $\text{top-}2$ matches defining the loop edge set~$\mathcal{E}_{\mathcal{L}_k}$, after which $z_k$ is appended for future queries. 
For each edge $(i,j)$ in the pose graph, the reprojection residual is defined as \donguk{similar form as} in~\eqref{eq:multi_view_residual}. 
The pose graph is then optimized by second-order IRLS solver, with its edge set incrementally expanded to include both sequential edges $\mathcal{E}_{\mathcal{S}_k}$ and loop edges $\mathcal{E}_{\mathcal{L}_k}$ for each keyframe $I_k$ as 
$
    \mathcal{E}_{\mathcal{G}} \leftarrow \mathcal{E}_{\mathcal{G}} \cup 
    (\mathcal{E}_{\mathcal{S}_k} \cup \mathcal{E}_{\mathcal{L}_k})
$. 
The corresponding keyframe set $\mathcal{W}_\text{pgo}$ contains $I_k$ and its connected neighbors. 
Unlike the local frontend subset $\mathcal{W}_\text{opt}$, the backend jointly aligns all sequential and loop edges in $\mathcal{E}_{\mathcal{G}}$ to enforce global consistency.

\section{Experimental Results and Discussion} \label{sec:exp_result}

%-------------------------------------------------------------------------
\subsection{Datasets and Evaluation Metrics}

We evaluate AIM-SLAM on the TUM RGB-D dataset~\cite{tum} and the EuRoC MAV dataset~\cite{euroc}. 
TUM RGB-D contains room-scale indoor trajectories with cluttered scenes. 
EuRoC features aggressive motions and large viewpoint changes, offering a challenging benchmark for robustness; in the uncalibrated setting on EuRoC, undistorted images are used for all methods without providing calibration, following~\cite{mast3r-slam}. 

Evaluation follows two standard metrics for dense visual SLAM: 
(i) camera pose estimation accuracy, measured by the RMSE of absolute trajectory error (ATE), and 
(ii) dense reconstruction quality, measured by accuracy, completion, and chamfer distance as similar to previous works~\cite{mast3r-slam, vggtslam, vggtlong}. 
For dense reconstruction evaluations on the EuRoC dataset, we only use the Vicon Room sequences where ground truth pointclouds were obtained from \donguk{laser} scans. 

%-------------------------------------------------------------------------
\subsection{Implementation Details and Baselines}

We used the released pretrained VGGT model with the same parameters across \donguk{all} datasets. 
The maximum size of the VGGT input subset~$\mathcal{W}$ is set to~$5$. 
All experiments are run on an NVIDIA RTX~3090 and Intel Core i9-11900K~(3.50~GHz), with input images resized to 518 pixels to meet VGGT requirements.

AIM-SLAM is designed for uncalibrated settings, and baselines are primarily evaluated under this condition. 
We compare AIM-SLAM against state-of-the-art learning-based \donguk{SLAMs}, including MASt3R-SLAM~\cite{mast3r-slam}, MUSt3R-VO~\cite{must3r}, VGGT-SLAM~\cite{vggtslam}, VGGT-Long~\cite{vggtlong}, and DROID-SLAM~\cite{droidslam}. 
All baselines are reported with loop closure enabled, except MUSt3R-VO, which does not support loop closure and is therefore evaluated in odometry mode. 
For DROID-SLAM, which assumes known intrinsics, we estimate intrinsics in the uncalibrated setting using GeoCalib~\cite{veicht2024geocalib} applied to the first frame of each sequence following prior works~\cite{vggtslam,vggtlong}.  
For completeness, we also report results in calibrated settings.

%-------------------------------------------------------------------------
\subsection{Analysis of Results}

%%%%%%%%%%%%%%%%%%%%%%%%%%%%%%%%%%%%%%%%%%%
\begin{table}[t!]
\begin{center}
\renewcommand{\arraystretch}{1.10} % height
\renewcommand{\tabcolsep}{0.35mm}  % width 
\caption{
    Quantitative comparison of camera pose accuracy on the TUM RGB-D dataset, measured by the RMSE of absolute trajectory error (ATE, unit: m). 
    We indicate the top three results as \colorbox{myFirst}{first}, \colorbox{mySecond}{second}, and \colorbox{myThird}{third}. 
}
\label{table:exp_main_pose_tum}
\resizebox{1.0\columnwidth}{!}
{
\mycustomsize
%%%%%%%%%%%%%%%%%%%
\begin{tabular}{lll@{\hspace{1mm}}*{9}{C{0.70cm}}C{0.70cm}}
\hline
                                                & \multicolumn{1}{c}{}                         &                         & \multicolumn{9}{c}{TUM RGB-D}                                                                                                                                                                                                                                                                                                                                                  &                                        \\ \cline{4-12}
\multirow{-2}{*}{}                              & \multicolumn{1}{c}{\multirow{-2}{*}{Method}} &                         & \texttt{360}                                    & \texttt{desk}                                   & \texttt{desk2}                                  & \texttt{floor}                                  & \texttt{plant}                                  & \texttt{room}                                   & \texttt{rpy}                                    & \texttt{teddy}                                  & \texttt{xyz}                                    & \multirow{-2}{*}{Avg.}                 \\ \hline
{\color[HTML]{C0C0C0} }                         & {\color[HTML]{C0C0C0} DeepV2D~\mbox{\cite{deepv2d}}}               & {\color[HTML]{C0C0C0} } & {\color[HTML]{C0C0C0} 0.243}           & {\color[HTML]{C0C0C0} 0.166}           & {\color[HTML]{C0C0C0} 0.379}           & {\color[HTML]{C0C0C0} 1.653}           & {\color[HTML]{C0C0C0} 0.203}           & {\color[HTML]{C0C0C0} 0.246}           & {\color[HTML]{C0C0C0} 0.105}           & {\color[HTML]{C0C0C0} 0.316}           & {\color[HTML]{C0C0C0} 0.064}           & {\color[HTML]{C0C0C0} 0.375}           \\
{\color[HTML]{C0C0C0} }                         & {\color[HTML]{C0C0C0} DeepFactors~\mbox{\cite{deepfactors}}}           & {\color[HTML]{C0C0C0} } & {\color[HTML]{C0C0C0} 0.159}           & {\color[HTML]{C0C0C0} 0.170}           & {\color[HTML]{C0C0C0} 0.253}           & {\color[HTML]{C0C0C0} 0.169}           & {\color[HTML]{C0C0C0} 0.305}           & {\color[HTML]{C0C0C0} 0.364}           & {\color[HTML]{C0C0C0} 0.043}           & {\color[HTML]{C0C0C0} 0.601}           & {\color[HTML]{C0C0C0} 0.035}           & {\color[HTML]{C0C0C0} 0.233}           \\
{\color[HTML]{C0C0C0} }                         & {\color[HTML]{C0C0C0} DROID-SLAM~\mbox{\cite{droidslam}}}            & {\color[HTML]{C0C0C0} } & {\color[HTML]{C0C0C0} 0.111}           & {\color[HTML]{C0C0C0} 0.018}           & {\color[HTML]{C0C0C0} 0.042}           & {\color[HTML]{C0C0C0} 0.021}           & {\color[HTML]{C0C0C0} 0.016}           & {\color[HTML]{C0C0C0} 0.049}           & {\color[HTML]{C0C0C0} 0.026}           & {\color[HTML]{C0C0C0} 0.048}           & {\color[HTML]{C0C0C0} 0.012}           & {\color[HTML]{C0C0C0} 0.038}           \\
{\color[HTML]{C0C0C0} }                         & {\color[HTML]{C0C0C0} DPV-SLAM~\mbox{\cite{dpvslam}}}              & {\color[HTML]{C0C0C0} } & {\color[HTML]{C0C0C0} 0.112}           & {\color[HTML]{C0C0C0} 0.018}           & {\color[HTML]{C0C0C0} 0.029}           & {\color[HTML]{C0C0C0} 0.057}           & {\color[HTML]{C0C0C0} 0.021}           & {\color[HTML]{C0C0C0} 0.330}           & {\color[HTML]{C0C0C0} 0.030}           & {\color[HTML]{C0C0C0} 0.084}           & {\color[HTML]{C0C0C0} 0.010}           & {\color[HTML]{C0C0C0} 0.076}           \\
\multirow{-5}{*}{{\color[HTML]{C0C0C0} \rot{Calib.}}} & {\color[HTML]{C0C0C0} MASt3R-SLAM~\mbox{\cite{mast3r-slam}}}           & {\color[HTML]{C0C0C0} } & {\color[HTML]{C0C0C0} 0.049}           & {\color[HTML]{C0C0C0} 0.016}           & {\color[HTML]{C0C0C0} 0.024}           & {\color[HTML]{C0C0C0} 0.025}           & {\color[HTML]{C0C0C0} 0.020}           & {\color[HTML]{C0C0C0} 0.061}           & {\color[HTML]{C0C0C0} 0.027}           & {\color[HTML]{C0C0C0} 0.041}           & {\color[HTML]{C0C0C0} 0.009}           & {\color[HTML]{C0C0C0} 0.030}           \\ \hline
                                                & DROID-SLAM~\mbox{\cite{droidslam}}                                   &                         & 0.202                                  & \cellcolor[HTML]{FFFFC7}0.032          & 0.091                                  & \cellcolor[HTML]{FFFFC7}0.064          & 0.045                                  & 0.918                                  & 0.056                                  & 0.045                                  & \cellcolor[HTML]{E9FFE2}0.012          & 0.158                                  \\
                                                & MUSt3R-VO~\mbox{\cite{must3r}}                                       &                         & 0.078                                  & 0.040                                  & \cellcolor[HTML]{FFFFC7}0.046          & 0.091                                  & 0.040                                  & \cellcolor[HTML]{E9FFE2}0.099          & 0.043                                  & \cellcolor[HTML]{FFFFC7}0.042          & \cellcolor[HTML]{FFFFC7}0.013          & \cellcolor[HTML]{FFFFC7}0.055          \\
                                                & VGGT-Long~\mbox{\cite{vggtlong}}                                    &                         & \cellcolor[HTML]{E9FFE2}0.053          & 0.064                                  & 0.060                                  & 0.111                                  & 0.064                                  & 0.170                                  & \cellcolor[HTML]{FFFFC7}0.036          & 0.127                                  & 0.047                                  & 0.081                                  \\
                                                & VGGT-SLAM~\mbox{\cite{vggtslam}}                                    &                         & 0.071                                  & \cellcolor[HTML]{E9FFE2}0.025          & \cellcolor[HTML]{E9FFE2}0.040          & 0.141                                  & \cellcolor[HTML]{C9FFB5}\textbf{0.023} & \cellcolor[HTML]{FFFFC7}0.102          & \cellcolor[HTML]{E9FFE2}0.030          & \cellcolor[HTML]{C9FFB5}\textbf{0.034} & 0.014                                  & \cellcolor[HTML]{E9FFE2}0.053          \\
                                                & MASt3R-SLAM~\mbox{\cite{mast3r-slam}}                                  &                         & \cellcolor[HTML]{FFFFC7}0.070          & 0.035                                  & 0.055                                  & \cellcolor[HTML]{E9FFE2}0.056          & \cellcolor[HTML]{FFFFC7}0.035          & 0.118                                  & 0.041                                  & 0.114                                  & 0.020                                  & 0.060                                  \\
\multirow{-6}{*}{\rot{Uncalib.}}                      & AIM-SLAM (ours)                             &                         & \cellcolor[HTML]{C9FFB5}\textbf{0.050} & \cellcolor[HTML]{C9FFB5}\textbf{0.017} & \cellcolor[HTML]{C9FFB5}\textbf{0.028} & \cellcolor[HTML]{C9FFB5}\textbf{0.024} & \cellcolor[HTML]{E9FFE2}0.026          & \cellcolor[HTML]{C9FFB5}\textbf{0.062} & \cellcolor[HTML]{C9FFB5}\textbf{0.021} & \cellcolor[HTML]{E9FFE2}0.039          & \cellcolor[HTML]{C9FFB5}\textbf{0.010} & \cellcolor[HTML]{C9FFB5}\textbf{0.031} \\ \hline
\end{tabular}
%%%%%%%%%%%%%%%%%%%
}
\vspace{-0.8cm}
\end{center}
\end{table}

%%%%%%%%%%%%%%%%%%%%%%%%%%%%%%%%%%%%%%%%%%%

%-------------------------------------------------------------------------
\subsubsection{Camera Pose Estimation}

% tum
{
    On the TUM RGB-D dataset, all methods achieve relatively high accuracy, as shown in Table~\ref{table:exp_main_pose_tum}. 
    AIM-SLAM stands out, surpassing the calibrated DROID-SLAM and achieving accuracy comparable with MASt3R-SLAM, \donguk{while requiring no camera intrinsics.}
}

%% euroc
{
    On the other hand, Table~\ref{table:exp_main_pose_euroc} reports the pose accuracy results on the EuRoC dataset. 
    % baselines
    For VGGT-Long and VGGT-SLAM, errors stem from submap-based alignment, where local predictions \donguk{of each submap} are reliable, but alignment fails under large viewpoint changes. 
    MASt3R-SLAM shows better robustness but remains limited by its two-view design, restricting the use of wide-baseline cues. 
    % ours
    In contrast, AIM-SLAM achieves the best overall accuracy among uncalibrated methods. 
    This highlights the effectiveness of the adaptive multi-view prioritization under challenging wide-baseline observations. 
}

%-------------------------------------------------------------------------
\subsubsection{Dense Reconstruction}\label{sec:exp_recon}

\begin{figure}[t]
    \centering
    \includegraphics[width=1.0\columnwidth, keepaspectratio]{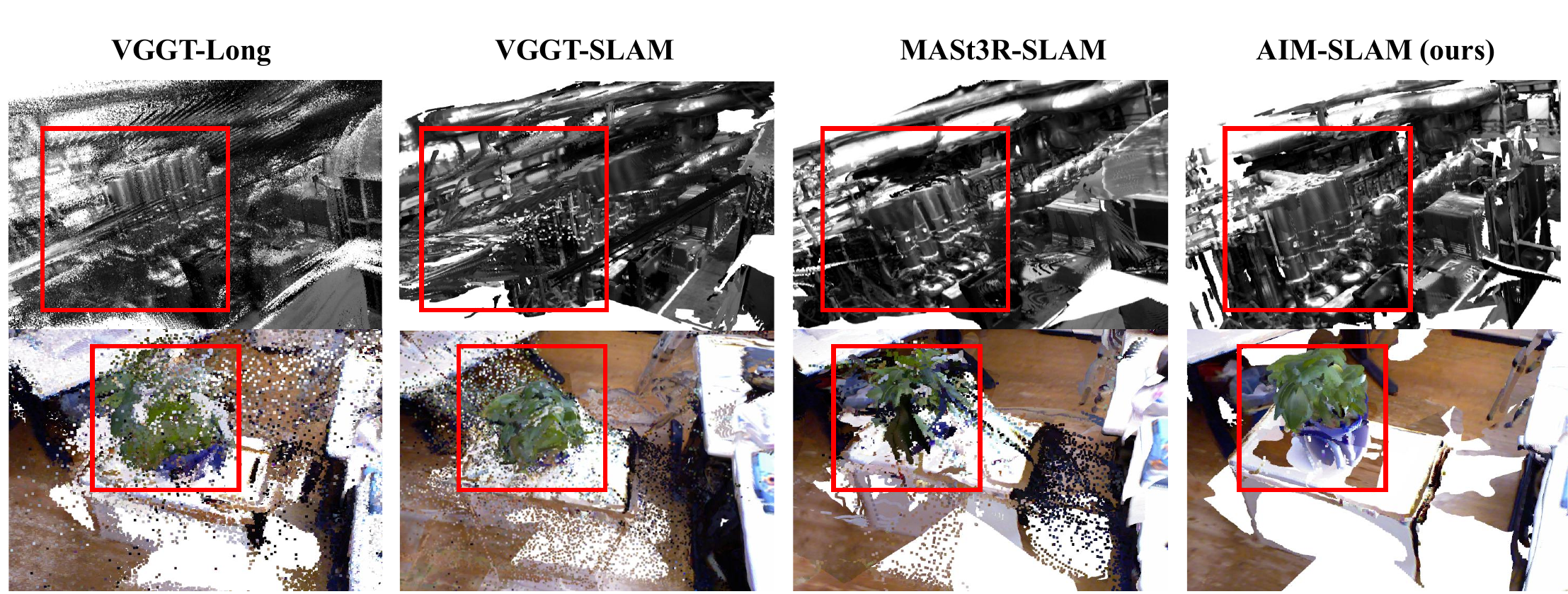}
    \caption{
        Dense reconstruction results on the EuRoC dataset (top row) and TUM RGB-D dataset (bottom row). 
        With its adaptive multi-view design, the proposed method achieves robust dense reconstructions across diverse environments. 
    }
    \label{fig:exp_recon_all}
\vspace{-0.4cm}
\end{figure}

%% TUM
{
    On the TUM RGB-D dataset, AIM-SLAM reconstructs fine object details more accurately than baselines, as shown in the bottom row of Fig.~\ref{fig:exp_recon_all}, \rev{achieving the strongest reconstruction performance across all metrics. }
}
%% euroc
{
    The same multi-view strategy also preserves global consistency in large-scale sequences, as demonstrated on the EuRoC dataset (top row of Fig.~\ref{fig:exp_recon_all}). 
    Here, baseline methods often suffer from ghosting artifacts on planar surfaces due to scale inconsistency, which persist even after two-view $\mathrm{Sim}(3)$ optimization~\cite{mast3r-slam} or accumulated submap alignment errors~\cite{vggtlong}. 
    \rev{
    Consistent with this observation, Table~\ref{table:exp_main_recon_all} shows that AIM-SLAM achieves the best accuracy and competitive completion and chamfer distance. 
    We attribute this behavior to a trade-off between stable informative-view selection and dense surface coverage, as AIM-SLAM focuses on a compact subset of reliable and overlapping views, which improves pose stability and geometric accuracy. }
}

%%%%%%%%%%%%%%%%%%%%%%%%%%%%%%%%%%%%%%%%%%%
\begin{table}[t!]
\begin{center}
\renewcommand{\arraystretch}{1.10} % height
\renewcommand{\tabcolsep}{0.35mm}  % width 
\caption{
    Quantitative comparison of camera pose accuracy on the EuRoC dataset, measured by the RMSE of absolute trajectory error (ATE, unit: m). 
    We indicate the top three results as \colorbox{myFirst}{first}, \colorbox{mySecond}{second}, and \colorbox{myThird}{third}. 
    \pf{$\dagger$~denotes that the average is computed excluding divergent sequences. }
}
\label{table:exp_main_pose_euroc}
\resizebox{1.0\columnwidth}{!}
{
\mycustomsize
%%%%%%%%%%%%%%%%%%%
% \begin{tabular}{lllcccccccccccc}
\begin{tabular}{lll@{\hspace{1mm}}*{11}{C{0.70cm}}C{0.70cm}}
\hline
                                                & \multicolumn{1}{c}{}                         &                         & \multicolumn{11}{c}{EuRoC}                                                                                                                                                                                                                                                                                                                                                                                                                                       &                                        \\ \cline{4-14}
\multirow{-2}{*}{}                              & \multicolumn{1}{c}{\multirow{-2}{*}{Method}} &                         & \texttt{V101}                                   & \texttt{V102}                                   & \texttt{V103}                                   & \texttt{V201}                                   & \texttt{V202}                                   & \texttt{V203}                                   & \texttt{MH01}                                   & \texttt{MH02}                                   & \texttt{MH03}                                   & \texttt{MH04}                                   & \texttt{MH05}                                   & \multirow{-2}{*}{Avg.}                 \\ \hline
{\color[HTML]{C0C0C0} }                         & {\color[HTML]{C0C0C0} DeepV2D~\mbox{\cite{deepv2d}}}               & {\color[HTML]{C0C0C0} } & {\color[HTML]{C0C0C0} 0.981}           & {\color[HTML]{C0C0C0} 0.801}           & {\color[HTML]{C0C0C0} 1.570}           & {\color[HTML]{C0C0C0} 0.290}           & {\color[HTML]{C0C0C0} 2.202}           & {\color[HTML]{C0C0C0} 2.743}           & {\color[HTML]{C0C0C0} 0.739}           & {\color[HTML]{C0C0C0} 1.144}           & {\color[HTML]{C0C0C0} 0.752}           & {\color[HTML]{C0C0C0} 1.492}           & {\color[HTML]{C0C0C0} 1.567}           & {\color[HTML]{C0C0C0} 1.298}           \\
{\color[HTML]{C0C0C0} }                         & {\color[HTML]{C0C0C0} DeepFactors~\mbox{\cite{deepfactors}}}           & {\color[HTML]{C0C0C0} } & {\color[HTML]{C0C0C0} 1.520}           & {\color[HTML]{C0C0C0} 0.679}           & {\color[HTML]{C0C0C0} 0.900}           & {\color[HTML]{C0C0C0} 0.876}           & {\color[HTML]{C0C0C0} 1.905}           & {\color[HTML]{C0C0C0} 1.021}           & {\color[HTML]{C0C0C0} 1.587}           & {\color[HTML]{C0C0C0} 1.479}           & {\color[HTML]{C0C0C0} 3.139}           & {\color[HTML]{C0C0C0} 5.331}           & {\color[HTML]{C0C0C0} 4.002}           & {\color[HTML]{C0C0C0} 2.040}           \\
{\color[HTML]{C0C0C0} }                         & {\color[HTML]{C0C0C0} DROID-SLAM~\mbox{\cite{droidslam}}}            & {\color[HTML]{C0C0C0} } & {\color[HTML]{C0C0C0} 0.037}           & {\color[HTML]{C0C0C0} 0.013}           & {\color[HTML]{C0C0C0} 0.019}           & {\color[HTML]{C0C0C0} 0.017}           & {\color[HTML]{C0C0C0} 0.010}           & {\color[HTML]{C0C0C0} 0.013}           & {\color[HTML]{C0C0C0} 0.013}           & {\color[HTML]{C0C0C0} 0.012}           & {\color[HTML]{C0C0C0} 0.022}           & {\color[HTML]{C0C0C0} 0.048}           & {\color[HTML]{C0C0C0} 0.044}           & {\color[HTML]{C0C0C0} 0.022}           \\
{\color[HTML]{C0C0C0} }                         & {\color[HTML]{C0C0C0} DPV-SLAM~\mbox{\cite{dpvslam}}}              & {\color[HTML]{C0C0C0} } & {\color[HTML]{C0C0C0} 0.035}           & {\color[HTML]{C0C0C0} 0.008}           & {\color[HTML]{C0C0C0} 0.015}           & {\color[HTML]{C0C0C0} 0.020}           & {\color[HTML]{C0C0C0} 0.011}           & {\color[HTML]{C0C0C0} 0.040}           & {\color[HTML]{C0C0C0} 0.013}           & {\color[HTML]{C0C0C0} 0.016}           & {\color[HTML]{C0C0C0} 0.022}           & {\color[HTML]{C0C0C0} 0.043}           & {\color[HTML]{C0C0C0} 0.041}           & {\color[HTML]{C0C0C0} 0.024}           \\
\multirow{-5}{*}{{\color[HTML]{C0C0C0} \rot{Calib.}}} & {\color[HTML]{C0C0C0} MASt3R-SLAM~\mbox{\cite{mast3r-slam}}}           & {\color[HTML]{C0C0C0} } & {\color[HTML]{C0C0C0} 0.040}           & {\color[HTML]{C0C0C0} 0.019}           & {\color[HTML]{C0C0C0} 0.027}           & {\color[HTML]{C0C0C0} 0.020}           & {\color[HTML]{C0C0C0} 0.025}           & {\color[HTML]{C0C0C0} 0.043}           & {\color[HTML]{C0C0C0} 0.023}           & {\color[HTML]{C0C0C0} 0.017}           & {\color[HTML]{C0C0C0} 0.057}           & {\color[HTML]{C0C0C0} 0.113}           & {\color[HTML]{C0C0C0} 0.067}           & {\color[HTML]{C0C0C0} 0.041}           \\ \hline
                                                & DROID-SLAM~\mbox{\cite{droidslam}}                                   &                         & 0.465                                  & 1.679                                  & 1.439                                  & 0.878                                  & 1.414                                  & 1.895                                  & \cellcolor[HTML]{E9FFE2}0.154          & \cellcolor[HTML]{FFFFC7}0.256          & 1.010                                  & 0.719                                  & 0.762                                  & 0.970                                  \\
                                                & MUSt3R-VO~\mbox{\cite{must3r}}                                    &                         & 0.489                                  & 0.287                                  & 0.554                                  & \cellcolor[HTML]{FFFFC7}0.123          & \cellcolor[HTML]{FFFFC7}0.109          & 0.252                                  & 0.265                                  & -                                      & 0.909                                  & 0.741                                  & 0.828                                  &  0.456$^{\dagger}$                                    \\
                                                & VGGT-Long~\mbox{\cite{vggtlong}}                                    &                         & 0.139                                  & \cellcolor[HTML]{FFFFC7}0.165          & \cellcolor[HTML]{FFFFC7}0.198          & 0.202                                  & 0.130                                  & \cellcolor[HTML]{E9FFE2}0.137          & 0.579                                  & 0.745                                  & \cellcolor[HTML]{FFFFC7}0.428          & \cellcolor[HTML]{FFFFC7}0.713          & \cellcolor[HTML]{FFFFC7}0.605          & \cellcolor[HTML]{FFFFC7}0.367          \\
                                                & VGGT-SLAM~\mbox{\cite{vggtslam}}                                    &                         & \cellcolor[HTML]{E9FFE2}0.098          & 0.184                                  & 0.353                                  & \cellcolor[HTML]{E9FFE2}0.068          & 0.903                                  & 0.431                                  & 0.400                                  & 0.701                                  & 3.599                                  & -                                      & -                                      & 0.749$^{\dagger}$                                    \\
                                                & MASt3R-SLAM~\mbox{\cite{mast3r-slam}}                                  &                         & \cellcolor[HTML]{FFFFC7}0.101          & \cellcolor[HTML]{E9FFE2}0.134          & \cellcolor[HTML]{E9FFE2}0.096          & 0.133                                  & \cellcolor[HTML]{E9FFE2}0.100          & \cellcolor[HTML]{FFFFC7}0.170          & \cellcolor[HTML]{FFFFC7}0.180          & \cellcolor[HTML]{E9FFE2}0.124          & \cellcolor[HTML]{E9FFE2}0.156          & \cellcolor[HTML]{E9FFE2}0.282          & \cellcolor[HTML]{E9FFE2}0.327          & \cellcolor[HTML]{E9FFE2}0.164          \\
\multirow{-6}{*}{\rot{Uncalib.}}                      & AIM-SLAM (ours)                             &                         & \cellcolor[HTML]{C9FFB5}\textbf{0.081} & \cellcolor[HTML]{C9FFB5}\textbf{0.059} & \cellcolor[HTML]{C9FFB5}\textbf{0.069} & \cellcolor[HTML]{C9FFB5}\textbf{0.057} & \cellcolor[HTML]{C9FFB5}\textbf{0.053} & \cellcolor[HTML]{C9FFB5}\textbf{0.060} & \cellcolor[HTML]{C9FFB5}\textbf{0.055} & \cellcolor[HTML]{C9FFB5}\textbf{0.076} & \cellcolor[HTML]{C9FFB5}\textbf{0.058} & \cellcolor[HTML]{C9FFB5}\textbf{0.115} & \cellcolor[HTML]{C9FFB5}\textbf{0.114} & \cellcolor[HTML]{C9FFB5}\textbf{0.072} \\ \hline
\end{tabular}
%%%%%%%%%%%%%%%%%%%
}
\vspace{-0.2cm}
\end{center}
\end{table}
%%%%%%%%%%%%%%%%%%%%%%%%%%%%%%%%%%%%%%%%%%%
%%%%%%%%%%%%%%%%%%%%%%%%%%%%%%%%%%%%%%%%%%%
\begin{table}[t!]
\begin{center}
\renewcommand{\arraystretch}{1.10} % height
\renewcommand{\tabcolsep}{0.35mm}  % width 
\caption{
    Quantitative comparison with state-of-the-art methods \pf{for} dense reconstruction on the EuRoC dataset (left) and TUM RGB-D dataset (right). 
    The best results are highlighted in \textbf{bold}, and the second-best results are \ul{underlined}. 
}
\label{table:exp_main_recon_all}
% \resizebox{1.0\columnwidth}{!}
{
\mycustomsize
%%%%%%%%%%%%%%%%%%%
\begin{tabular}{llccccccc}
\hline
\multicolumn{1}{c}{\multirow{2}{*}{Method}} &  & \multicolumn{3}{c}{EuRoC}                                     &  & \multicolumn{3}{c}{TUM RGB-D}                                 \\ \cline{3-5} \cline{7-9} 
\multicolumn{1}{c}{}                        &  & Accuracy       & Completion     & \multicolumn{1}{l}{Chamfer} &  & Accuracy       & Completion     & \multicolumn{1}{l}{Chamfer} \\ \cline{1-1} \cline{3-5} \cline{7-9} 
VGGT-Long~\mbox{\cite{vggtlong}}                                   &  & \ul{0.106}          & 0.119 & 0.112                 &  & \ul{0.094}          & \ul{0.107}          & \ul{0.100}                       \\
VGGT-SLAM~\mbox{\cite{vggtslam}}                                   &  & 0.246          & 0.216          & 0.231                       &  & 0.109    & 0.127    & 0.118                 \\
MASt3R-SLAM~\mbox{\cite{mast3r-slam}}                                 &  & 0.108    & \textbf{0.072}          & \textbf{0.090 }                      &  & 0.097          & 0.113          & 0.105                       \\
AIM-SLAM (ours)                            &  & \textbf{0.103} & \ul{0.102}    & \ul{0.102}              &  & \textbf{0.063} & \textbf{0.098} & \textbf{0.081}              \\ \hline
\end{tabular}
%%%%%%%%%%%%%%%%%%%
}
\vspace{-0.7cm}
\end{center}
\end{table}
%%%%%%%%%%%%%%%%%%%%%%%%%%%%%%%%%%%%%%%%%%%

%%%%%%%%%%%%%%%%%%%%%%
%%%%%%%%%%%%%%%%%%%%%%

%%%%%%%%%%%%%%%%%%%%%%
\begin{figure}[t!]
    \centering
    \subfigure[]{\includegraphics[width=0.4\linewidth]{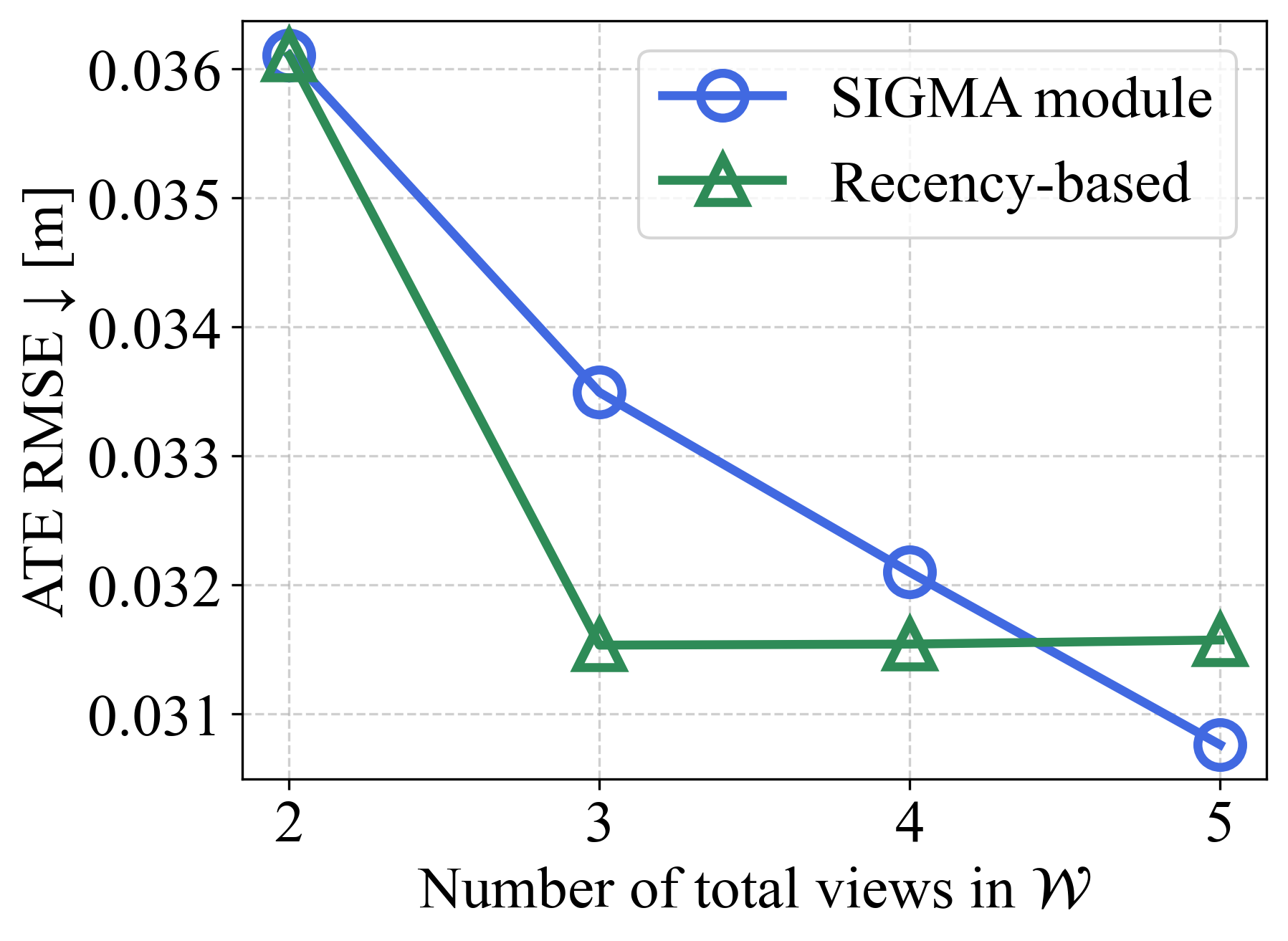}\label{fig:view_rmse_tum}}
    \subfigure[]{\includegraphics[width=0.4\linewidth]{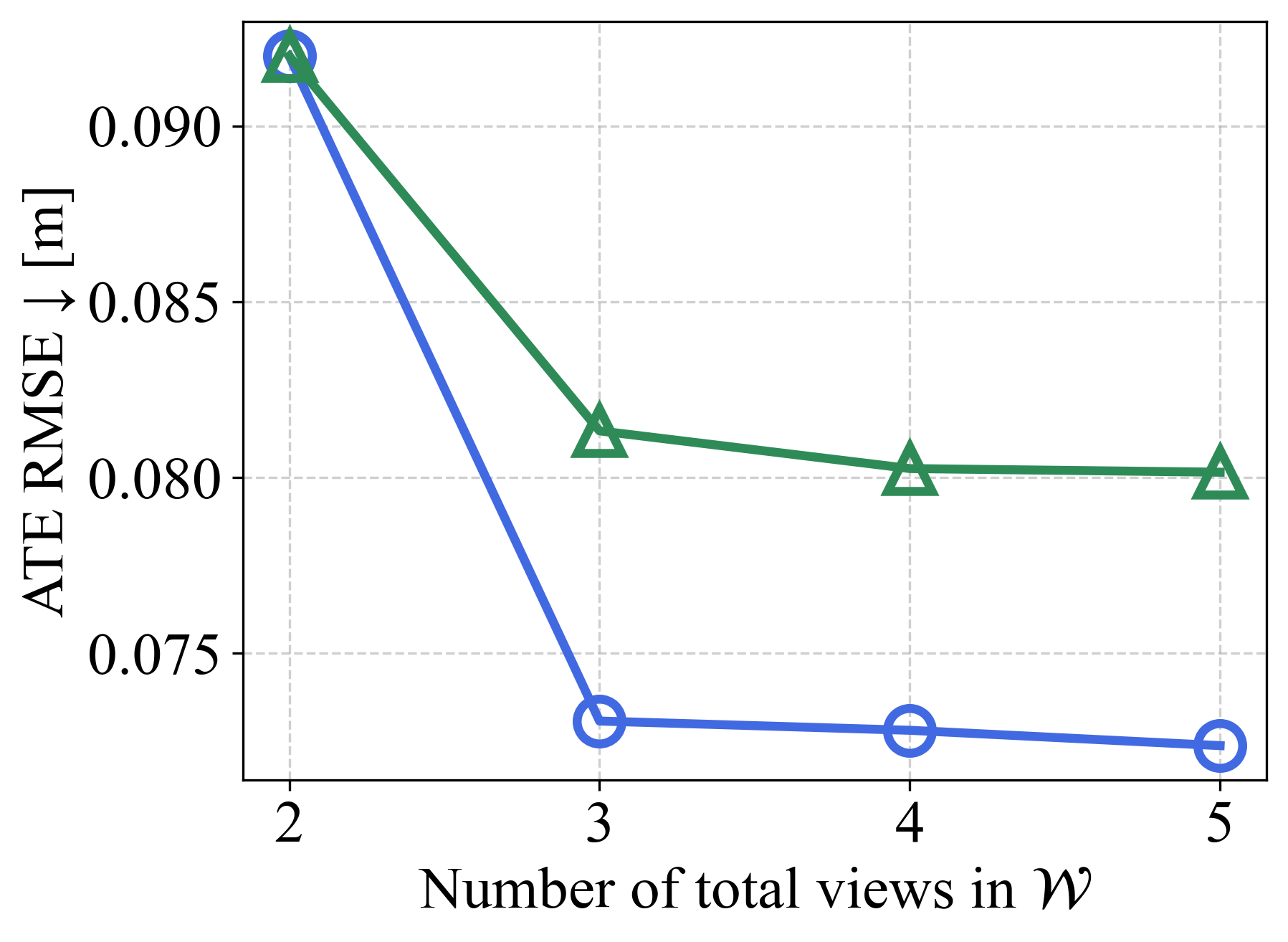}\label{fig:view_rmse_euroc}}
    \vspace{-0.2cm}
    \caption{
        Ablation study of the proposed SIGMA module showing how the total number of views affects pose accuracy (ATE RMSE) for the 
        \fn{
        \subref{fig:view_rmse_tum}~TUM RGB-D dataset and 
        \subref{fig:view_rmse_euroc}~EuRoC dataset. 
        }
        Recency-based denotes selecting the most recent consecutive keyframes to form the input subset $\mathcal{W}$, as in prior methods~\cite{mast3r-slam, vggtslam, vggtlong}. 
    }
    \label{fig:exp_abl_numview}
% \vspace{\vsfig}
\end{figure}
%%%%%%%%%%%%%%%%%%%%%%

\subsection{Ablation Study}\label{sec:exp_abl}

% We conduct ablation studies to analyze the contribution of each design choice in PRiSM-SLAM. 

\subsubsection{Effect of the \pf{Total Number of Views}}

Fig.~\ref{fig:exp_abl_numview} compares pose accuracy as the maximum limit size of the input keyframe subset~$\mathcal{W}$ increases, between the cases where~$\mathcal{W}$ is constructed in a recency-based manner and the case where the SIGMA module is employed. 
For the 2-view and 3-view cases, the subset size is fixed without adaptive expansion for the comparison. 
Increasing the limit naturally improves accuracy by introducing additional multi-view constraints, but the gain quickly saturates beyond 4–5 views. 
On the TUM dataset, both methods achieve comparable performance. 
The difference between the two methods becomes evident on the EuRoC dataset, which involves larger baselines and rapid viewpoint changes \fn{(Fig.~\ref{fig:view_rmse_euroc})}. 
While both methods eventually saturate beyond 4–5 views, the SIGMA module-based method maintains substantially higher accuracy throughout, as it consistently leverages more informative keyframes than the recency-based strategy.

%%%%

\subsubsection{Effect of the \pf{Hybrid Residual}}

Table~\ref{table:exp_abl_hybrid} compares alternative residual formulations.
Ray-only residuals result in the largest errors, showing that the lack of pixel-level constraints limits geometric precision. 
Projection-only residuals with VGGT-estimated intrinsics reduce error but remain sensitive to calibration noise.
The hybrid formulation yields the best performance on both datasets by combining the angular robustness of rays with the pixel-level accuracy of projections, highlighting the complementarity of the two terms.

%%%%%%%%%%%%%%%%%%%%%%%%%%%%%%%%%%%%%%%%%%%
\begin{table}[t!]
\begin{center}
\renewcommand{\arraystretch}{1.10} % height
\renewcommand{\tabcolsep}{0.35mm}  % width 
\caption{
    Ablation study of the proposed hybrid residual for joint multi-view $\mathrm{Sim}(3)$ pose optimization on the EuRoC dataset (left) and the TUM RGB-D dataset (right).
}
\label{table:exp_abl_hybrid}
% \resizebox{1.0\columnwidth}{!}
{
\mycustomsize
%%%%%%%%%%%%%%%%%%%
\begin{tabular}{llcc}
\hline
\multicolumn{1}{c}{\multirow{2}{*}{Method}} &  & \multicolumn{2}{c}{ATE RMSE}    \\ \cline{3-4} 
\multicolumn{1}{c}{}                        &  & EuRoC          & TUM RGB-D      \\ \cline{1-1} \cline{3-4} 
Ray only                                 &  & 0.138          & 0.061          \\
Projection only                              &  & 0.081          & 0.032          \\
Hybrid (ray + projection)                 &  & \textbf{0.072} & \textbf{0.031} \\ \hline
\end{tabular}
%%%%%%%%%%%%%%%%%%%
}
\vspace{-0.8cm}
\end{center}
\end{table}
%%%%%%%%%%%%%%%%%%%%%%%%%%%%%%%%%%%%%%%%%%%

\section{Conclusions} \label{sec:cons}

% summary
In summary, we presented AIM-SLAM, a dense monocular SLAM framework for uncalibrated settings that leverages geometry-aware foundation models. 
\donguk{The proposed SIGMA module} adaptively \pf{prioritizes} a sparse but overlap-rich and informative keyframe subset. 
The prioritized multi-view subset is then jointly optimized in $\mathrm{Sim}(3)$ space to reduce both short- and mid-term drift. 
Together, these components enable accurate \rev{pose estimation} and \rev{geometrically} consistent dense reconstruction under uncalibrated conditions, offering a more scalable solution for foundation model-based SLAM.
% limitation
The current limitation of AIM-SLAM is its reliance on VGGT inference, yielding an overall runtime of about 3~Hz in our environment. 
Excluding VGGT inference, the remaining components of our method run at about 17~Hz. Future work will explore accelerating the current foundation model or integrating faster alternatives.

%%%%%%%%%%%%%%%%%%%%%%%%%%%%%%%%%%%%%%%%%%%%%%%%%%%%%%%%%%%%%%%%%%%%%%%%%%%%%%%%

% \newpage

\bibliographystyle{IEEEtran}
% \bibliography{ref_alias_long, references}

\bibliography{ref_alias_short, references}

\end{document}